\pdfoutput=1

\documentclass[11pt]{article}

\usepackage{acl}

\usepackage{times}
\usepackage{latexsym}
\usepackage{booktabs}
\usepackage{array}
\usepackage{soul}
\usepackage{amsmath}
\usepackage{graphicx} 
\usepackage{xspace}
\usepackage{caption}
\usepackage{subcaption}
\usepackage{amsfonts}
\usepackage{multirow}
\usepackage{color}
\usepackage[shortlabels]{enumitem}
\usepackage{pifont}

\usepackage{stfloats}
\usepackage{mathtools,algorithm}
\usepackage[noend]{algpseudocode}

\usepackage[T1]{fontenc}

\usepackage[utf8]{inputenc}

\usepackage{microtype}

\makeatletter
\DeclareRobustCommand\onedot{\futurelet\@let@token\@onedot}
\def\@onedot{\ifx\@let@token.\else.\null\fi\xspace}

\DeclareRobustCommand\rightbracket{\futurelet\@let@token\@rightbracket}
\def\@rightbracket{\ifx\@let@token]\else]\null\fi\xspace}
\def\mask{\textsc{[mask}\rightbracket}
\newcommand{\s}{\textcolor{white}{0}}

\def\Masking{Deterministic Masking\xspace}
\def\TaskA{Clue Contrastive Learning\xspace}

\def\taska{clue contrastive learning\xspace}

\def\TaskB{Clue Classification\xspace}

\def\taskb{clue classification\xspace}

\def\masking{deterministic masking\xspace}

\title{Pre-training Language Models with Deterministic Factual Knowledge}

\author{
Shaobo Li$^{1}$,
Xiaoguang Li$^{2}$,
Lifeng Shang$^{2}$,\\
\textbf{Chengjie Sun$^{1}$, Bingquan Liu$^{1}$, Zhenzhou Ji$^{1}$, Xin Jiang$^{2}$ and  Qun Liu$^{2}$}\\
$^{1}$Harbin Institute of Technology, 
$^{2}$Huawei Noah's Ark Lab\\
{shli@insun.hit.edu.cn},
{\{sunchengjie, liubq, jizhenzhou\}@hit.edu.cn}\\
{\{lixiaoguang11, shang.lifeng, Jiang.Xin, qun.liu\}@huawei.com}\\
}

\begin{document}
\maketitle
\begin{abstract}
Previous works show that Pre-trained Language Models (PLMs) can capture factual knowledge. However, some analyses reveal that PLMs fail to perform it robustly, e.g., being sensitive to the changes of prompts when extracting factual knowledge. To mitigate this issue, we propose to let PLMs learn the deterministic relationship between the remaining context and the masked content. The deterministic relationship ensures that the masked factual content can be deterministically inferable based on the existing clues in the context. That would provide more stable patterns for PLMs to capture factual knowledge than randomly masking. Two pre-training tasks are further introduced to motivate PLMs to rely on the deterministic relationship when filling masks. Specifically, we use an external Knowledge Base (KB) to identify deterministic relationships and continuously pre-train PLMs with the proposed methods. The factual knowledge probing experiments indicate that the continuously pre-trained PLMs achieve better robustness in factual knowledge capturing. Further experiments on question-answering datasets show that trying to learn a deterministic relationship with the proposed methods can also help other knowledge-intensive tasks.

\end{abstract}

\section{Introduction}

\citet{lama,JiangXAN20,shin2020autoprompt,optim-prompt} show that we can successfully extract factual knowledge from Pre-trained Language Models~(PLMs) using cloze-style prompts such as ``The director of the film Saving Private Ryan is \mask.'' Some recent works \cite{lanka,DBLP:e-BERT} find that the PLMs may rely on superficial cues to achieve that and can not respond robustly. Table~\ref{tab:intro} gives examples of inconsistent predictions exposed by changing the surface forms of prompts on the same fact.

This phenomenon questions whether PLMs can robustly capture factual knowledge through Masked Language Modeling~(MLM)~\cite{BERT} and further intensify us to inspect the masked contents in the pre-training samples. After reviewing several masking methods, we find that they focus on limiting the granularity of masked contents, e.g., restricting the masked content to be entities and then randomly masking the entities~\cite{SSM}, and pay less attention to checking whether the obtained MLM samples are appropriate for factual knowledge capturing. For instance, when we want PLMs to capture the corresponding factual knowledge as recovering the masked entities, we should check whether the remaining context provides sufficient clues to recover the missing entity.

\begin{table}[t]
    \centering
    \small
    \begin{tabular}{
    >{\arraybackslash}m{0.33\textwidth}
    c
    }
    \toprule
    \textbf{Cloze-style Prompt and \underline{Prediction}}    & \textbf{Is Correct?}            \\
    \midrule
    
     War Horse is an American war film directed by \underline{Steven Spielberg}.          & \ding{51}        \\
     The director of the American war film War Horse is \underline{Keanu Reeves}.         & \ding{55}    \\
    \underline{Christopher Nolan} is the director of the American war film War Horse.     & \ding{55}    \\
    \bottomrule
    \end{tabular}
    
    \caption{A PLM could gives inconsistent results when probing the same fact with different prompts. The underlined words are the predictions.}
    \label{tab:intro}
\end{table}

Inspired by the above analysis, we can categorize MLM samples based on the relationship between the remaining context and masked content:
\begin{itemize}
[itemsep=1pt,topsep=1pt,parsep=2pt]
    \item \textbf{Non-deterministic samples} The clues in the remaining context are insufficient to constrain the value of the masked content. Multiple values are valid to fill in the masks.
    \item \textbf{Deterministic samples} The remaining context holds deterministic clues for the masked content. We can get one and only one valid value for the masked content.
\end{itemize}

For example, the first cloze in Table~\ref{tab:intro} masks the director of the film ``\textit{War Horse}.'' Since the film has only one director in the real world, we can get a unique answer deterministically. So it is a deterministic MLM sample. The crucial clues ``\textit{War Horse}''  and ``\textit{directed by}'' have a deterministic relationship with the missing entity ``\textit{Steven Spielberg}.'' For brevity, we refer to these clues as \textbf{deterministic clues} and the outcome ``\textit{Steven Spielberg}" as \textbf{deterministic span}. In contrast, if the sample becomes ``\mask{}s \textit{is an American war film directed by Steven Spielberg},''  multiple names can fill the masks because Steven Spielberg produced more than one American war film. We cannot tell which one is better based on the existing clues, so it is a non-deterministic sample.

The non-deterministic samples establish a multi-label problem \cite{multi-label-def} for MLM, where more than one ground-truth value for outputs is associated with a single input. If we enforce the PLMs to promote one specified ground truth over others, the other ground truths become false negatives that could plague the training or cause a performance downgrade \cite{multi-label-missing-multi-labels,multi-label-false-negative}.
The non-deterministic samples are competent for obtaining contextualized representations but become questionable for understanding the intrinsic relationship between factual entities.
In contrast, the deterministic samples are less confusing since the answer is always unique, providing a stable relationship for PLMs to learn.

Therefore, we propose \textbf{\masking} that always masks and predicts the deterministic spans in MLM pre-training to improve PLMs' ability to capture factual knowledge. The deterministic clues and spans are identified based on a KB. Two pre-training tasks, \textbf{\taska} and \textbf{\taskb}, are introduced to make PLMs more aware of the deterministic clues when predicting the missing entities. The \taska encourages PLMs to be more confident in prediction~\cite{entropy-as-certainty-1,entropy-as-certainty-2} when the deterministic clues are unmasked. The \taskb is to detect whether the remaining context contains deterministic clues. The experiments on the factual knowledge probing and question-answering tasks show the effectiveness of the proposed methods.

The contributions of this paper are: (1) We propose to model the deterministic relationship in MLM samples to improve the robustness (i.e., both consistency and accuracy) of factual knowledge capturing. (2) We design two pre-training tasks to enhance the deterministic relationship between entities to earn further improvement on robustness. (3) The experiment results show that learning the deterministic relationship is also helpful for other knowledge-intensive tasks, such as question answering.

\section{Methods}

Section~\ref{sec:df-masking} expatiates the \masking, which includes how we align texts with triplets and identify deterministic clues and spans in texts. 
The \taska and \taskb are described in Sections~\ref{sec:task-cons} and \ref{sec:task-cls}, respectively.

\subsection{\Masking}
\label{sec:df-masking}

In addition to masking only factual content, the \masking also constrains the remaining context and the masked content to have a deterministic relationship: the remaining context should provide conclusive clues to predict the masked content, and the valid value to fill in the mask is unique.

To this end, we align each text with a KB triplet and match the spans in the text with (\textit{subject}, \textit{predicate}, \textit{object}) respectively. We select the spans aligned with \textit{object}s as the candidates to be masked for pre-training. To further make the masked object deterministic, we query the KB with the aligned (\textit{subject}, \textit{predicate}) and check whether the valid object that exists in KB is unique.

If the KB emits this object exclusively, e.g., only the aligned object can compose a valid triplet with the aligned \textit{subject} and \textit{predicate}, the object is deterministic. The object is non-deterministic if multiple objects suit the aligned subject and predicate in the KB. The span aligned with the deterministic object is a deterministic span, and it would be masked to construct a deterministic MLM sample\footnotemark. We pre-train PLMs on only the deterministic samples.
\footnotetext{We put the detailed procedure (includes entity linking and predicate string matching) in Appendix~\ref{sec:app_pt_data_cons}.}

\begin{figure}[ht]
    \centering
    \includegraphics[width=\columnwidth]{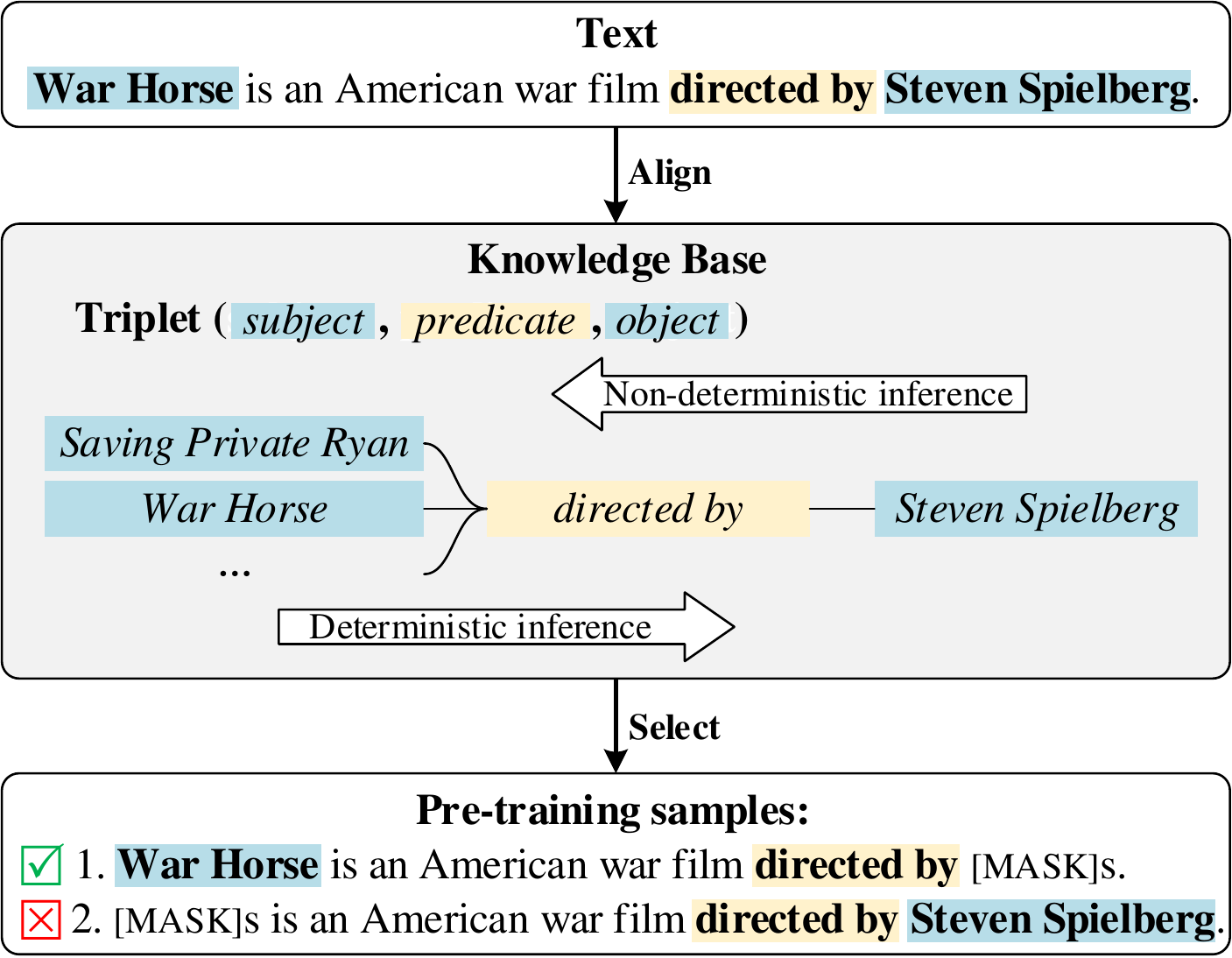}
    \caption{Construct a deterministic sample. The spans with blue background correspond to entities (subject or object), and the spans with yellow describe relations (predicate).}
    \label{fig:df-masking}
\end{figure}

Figure~\ref{fig:df-masking} shows a deterministic sample aligned with the triplet (``\textit{War Horse},'' ``\textit{directed by},'' ``\textit{Steven Spielberg}''). When querying KB with ``\textit{War Horse}'' as the subject and ``\textit{directed by}'' as the predicate, the result object ``\textit{Steven Spielberg}'' is unique because there is only one director who produced this film, so the first sample is deterministic. In contrast, when using ``\textit{Steven Spielberg}'' and ``\textit{directer of}'' as the subject and the predicate, multiple valid objects exist in KB, so the second sample is non-deterministic and is filtered out.

By dropping the non-deterministic samples, we prevent PLMs from having a crush on one object but ignoring others that are also valid based on the existing clues. While in the deterministic samples, the relationship between the remaining clues and the missing span is more stable and unambiguous. Training on the deterministic samples encourages PLM to infer the missing object based on its deterministic factual clues. It helps PLMs grasp a more substantial relationship between entities to model the factual contents and could aid in accomplishing some knowledge-intensive tasks.

\subsection{\TaskA}
\label{sec:task-cons}
\begin{figure}[ht]
    \centering
    \begin{subfigure}[b]{\columnwidth}
        \centering
         \includegraphics[width=\columnwidth]{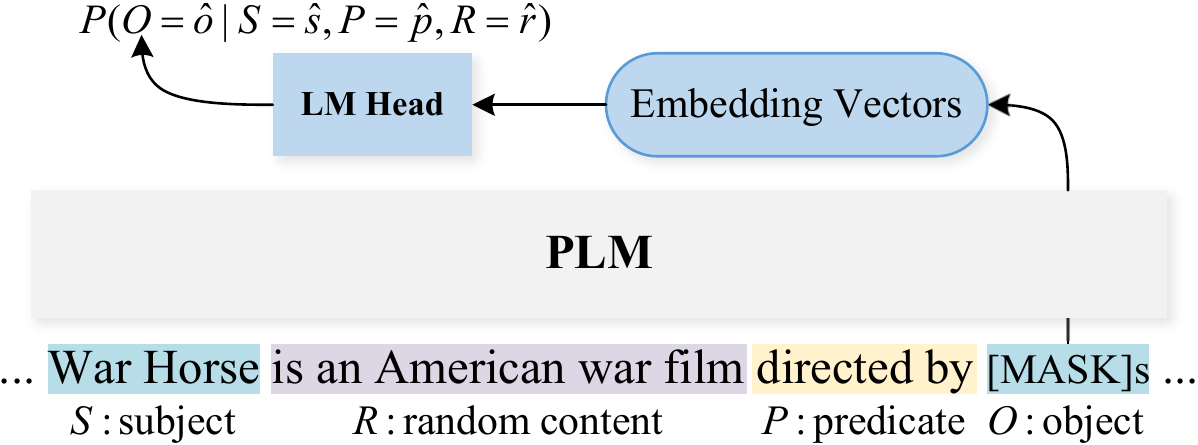}
         \caption{Deterministic sample: masks the deterministic span (object) and keeps the deterministic clues (object and predicate).}
         \label{fig:contrastive-ori}
    \end{subfigure}
    \par\bigskip
    \begin{subfigure}[b]{\columnwidth}
        \centering
        \includegraphics[width=\columnwidth]{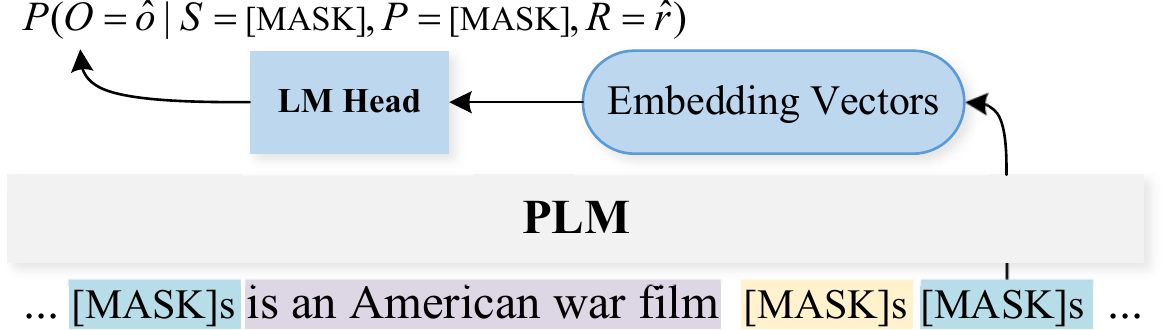}
        \caption{Contrastive sample without deterministic clues: masks both the deterministic clues and the deterministic span.}
        \label{fig:contrastive-dm}
    \end{subfigure}
    \caption{The two samples in \taska. The first sample (a) has a more informative context, so PLM should be more confident when predicting the masked object $O$. The texts with purple background denote the spans other than entities and relations.}
    \label{fig:contrastive-obj}
\end{figure}

To stimulate PLMs to catch the deterministic relationship between entities, we design the pre-training task \taska following this intuition: PLMs should have more confidence to generate a masked span when its deterministic clues exist in the context, and introduce a contrastive objective accordingly. We explain it with a pair of samples in Figure~\ref{fig:contrastive-obj}. Figure~\ref{fig:contrastive-ori} shows a deterministic MLM sample that masks the span ``\textit{Steven Spielberg}'' and keeps its deterministic clues. Figure~\ref{fig:contrastive-dm} masks both the deterministic clues and the deterministic span.
The remaining context in Figure~\ref{fig:contrastive-ori} contains fewer \mask{}s and provides more information, naturally reducing the uncertainty in prediction. So PLMs should assign a higher probability for the ground truth when giving the context in Figure~\ref{fig:contrastive-ori} than Figure~\ref{fig:contrastive-dm}.

Formally, we use $S$ and $P$ to denote the deterministic clues (subject and predicate) and $O$ to denote the masked deterministic span (object). $R$ represents the random spans in the context other than $S$, $P$, and $O$. The objective function that needs to be maximized is:
\begin{equation}
    \small
    \begin{split}
         &P(O=\hat{o}\mid S=\hat{s},P=\hat{p},R=\hat{r}) \\
        -&P(O=\hat{o}\mid S=\mask,P=\mask,R=\hat{r}),
    \end{split}
\end{equation}
$S=\mask$ and $P=\mask$ denote replacing the deterministic clues with \mask{}s. $\hat{s}$, $\hat{p}$, $\hat{o}$ and $\hat{r}$ are the ground-truth values of the $S$, $P$, $O$ and $R$, respectively.
$P(O=\hat{o}\mid \cdot)$ denotes the probability that the PLM correctly predicts the masked span $O$, i.e., the average probability that the PLM assigns to the ground-truth tokens. It is calculated by a Language Model Head (LM Head) based on the embedding of $O$ from the PLMs.

This task encourages PLMs to give the ground truth $\hat{o}$ a higher probability when the deterministic clues exist in the context. It is somewhat conservative since we consider the noise in the data construction. The objective is still reasonable even when the $S$, $P$, and $R$ are randomly labeled. Raw words are always more informative than the ordinary \mask{}s and can reduce the uncertainty of the context \cite{information-theory}, so the uncertainty of prediction degrades accordingly~\cite{entropy-as-certainty-1,entropy-as-certainty-2}. On the other hand, this objective trains PLMs to react to the changes in the context, i.e., learning how to tune the output as the input changes. We employ a large-scale KB as the approximation of real-world knowledge \cite{closed-world-assumption} to get the pre-training samples.

\subsection{\TaskB}
\label{sec:task-cls}
\begin{figure}[ht]
    \centering
        \centering
        \includegraphics[width=\columnwidth]{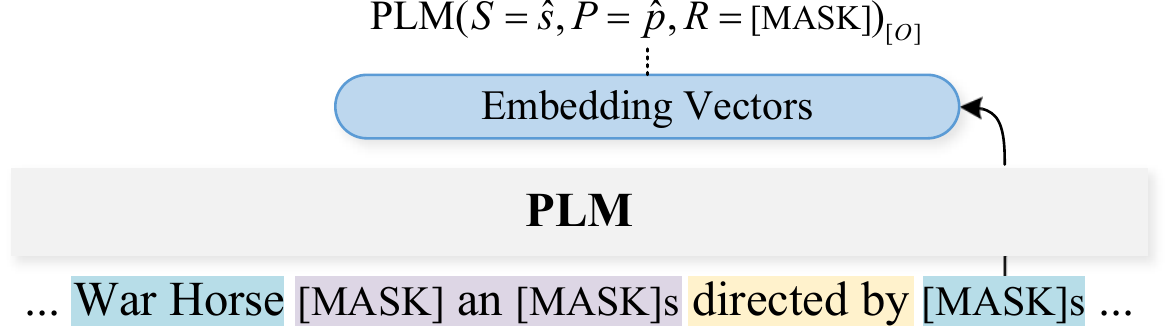}
        \caption{A sample that masks some random spans $R$ (with purple background) in the context.}
    
    \label{fig:clasification-obj}
\end{figure}

The \taskb asks PLMs to classify what kinds of clues exist in the remaining context. After masking the deterministic span $O$, we manipulate the remaining context to generate three samples that contain different kinds of contexts:

\begin{enumerate}[label=(\alph*),itemsep=1pt,topsep=1pt,parsep=2pt]
    \item \textbf{Keep deterministic clues:} we only mask the deterministic span $O$ and leave its deterministic clues untorched. It is the same as the original deterministic MLM sample shown in Figure~\ref{fig:contrastive-ori}.
    \label{sample:a}
    
    \item \textbf{Mask deterministic clues:} we mask $O$ and its deterministic clues ($S$ and $P$). It is the same as the constructive sample in Figure~\ref{fig:contrastive-dm}.
    \label{sample:b}
    
    \item \textbf{Mask random spans:} we mask $O$ and some random spans $R$ other than the deterministic clues. An example is shown in Figure~\ref{fig:clasification-obj}. The number of tokens in $R$ is the same as the number of tokens in the deterministic clues.
    \label{sample:c}
\end{enumerate}

The PLMs that use the Transformer encoder \cite{transformer} as the backbone emit a contextualized embedding for each input token. Every contextualized embedding can encode all the information in context since all the input tokens are involved when computing the embedding. So we encode the clues in the remaining context using the contextualized embedding of $O$. Formally, the clue representations for the above three samples are:

\begin{equation}
    \small
    \begin{split}
        E_{(a)} &= {\rm PLM}(S=\hat{s},P=\hat{p},R=\hat{r})_{[O]}, \\
        E_{(b)} &= {\rm PLM}(S=\mask,P=\mask,R=\hat{r})_{[O]}, \\
        E_{(c)} &= {\rm PLM}(S=\hat{s},P=\hat{p},R=\mask)_{[O]}, \\
        E_{(a)}&,E_{(b)},E_{(c)}\in\mathbb{R}^{\mid O\mid\times d}.
    \end{split}
\end{equation}
$E_{(a)}$, $E_{(b)}$, and $E_{(c)}$ represent the inputs that (a) keep deterministic clues $S$ and $P$, (b) mask deterministic clues, and (c) mask random spans $R$, respectively. ${\rm PLM}(\cdot)$ denotes the PLM that can output the contextualized embedding for each input token. ${\rm PLM}(\cdot)_{[O]}$ denotes grabbing the embedding vectors corresponding to $O$ from the PLM's output. Since the three input samples have the same $O$, the number of token-level embedding vectors is the same in $E_{(a)}$, $E_{(b)}$, and $E_{(c)}$.

Each token-level embedding vector $\mathbf{e}$ in $E_{(a)}$, $E_{(b)}$, and $E_{(c)}$ is fed into a three-way classifier:
\begin{equation}
\small
    y = {\rm{softmax}}(\mathbf{W}^\top\mathbf{e}), \mathbf{e}\in\mathbb{R}^{d}, \mathbf{W}\in\mathbb{R}^{d\times 3}.
\end{equation}
$y$ is a three-element vector and shows the probabilities that $\mathbf{e}$ comes from $E_{(a)}$, $E_{(b)}$ and $E_{(c)}$. $\mathbf{W}^\top$ is the three-way classifier.

The number of masked tokens in samples  \ref{sample:a} and \ref{sample:b} is different since the latter masks the clues additionally. It may become a shortcut for the proposed contrastive or classification tasks. So we introduce sample \ref{sample:c}, which has the same number of masks as sample \ref{sample:b}, to eliminate this shortcut. 
Some existing pre-training methods \cite{electra,WKLM, kgplm} replace original tokens with fake tokens to build the pseudo samples and then classify between original and pseudo samples, while \taskb employs \mask{}s in the replacement. We wonder that fake tokens may make the intervened input tell fake facts conflict with the real world, leading the PLMs to capture wrong knowledge from the pseudo samples. \mask{} is a safer choice here. 

\section{Experiments}

\subsection{Pre-training Data}
\label{sec:exp_data_construction}
We use Wikipedia\footnote{https://www.wikipedia.org/} as the source of texts and Wikidata\footnote{https://www.wikidata.org/} as the knowledge base. We split the Wikipedia texts into natural paragraphs and then align each paragraph with subject-predicate-object triplets in WikiData. Each aligned object that is deterministic based on WikiData is the deterministic span, and all the subjects and predicates that correspond to the deterministic span are deterministic clues. The paragraphs with identified deterministic spans and clues are then used for pre-training.

We first employ the TREX~\cite{TREX} that provides the alignments between texts and triplets to construct a preliminary dataset named \textbf{Partial data}. About 46.1\% of triplets are non-deterministic and ignored in the partial data.
TERX provides aligned triplets for only the abstract paragraphs (first paragraph) in Wikipedia. We enlarge the data size by processing all the paragraphs in Wikipedia. The detailed process is in Appendix~\ref{sec:app_pt_data_cons}. The dataset that involves all the paragraphs is referred to as \textbf{Full data}. Table~\ref{tab:pre-training-data} shows the statistics of the two pre-training datasets. For efficiency reasons, we use the partial data to train the baselines in the ablation study. The full data is for our final model. 

\begin{table}[htbp]
    \small
    \centering
    \begin{tabular}{lcc}
    \toprule
    {}                                  & \textbf{Partial data} & \textbf{Full data} \\
    \midrule

    \# Paragraphs                        & 3,726,205     & 17,373,859 \\
    \# Samples                           & 12,134,717    & 39,289,518 \\
    Avg. tokens per paragraph            & 166.93        & 134.61     \\
    Avg. \# tokens in per $S \cup P$     & 8.53          & 7.52       \\
    Avg. \# tokens in per $O$            & 3.19          & 2.99       \\
    \bottomrule
    \end{tabular}
    \caption{Statistics of the pre-training data.}
    \label{tab:pre-training-data}

\end{table}

\subsection{Evaluation Tasks and Datasets}
\label{sec:evaluation-tasks}
We first evaluate the proposed method with \textit{cloze-style QA}. Then We adopt two other knowledge-intensive tasks, \textit{closed-book QA} and \textit{extractive QA}, to evaluate the PLMs' ability to capture and understand factual knowledge.

\subsubsection{Cloze-style QA}
\label{sec:task-cloze-style-qa}
Following \citet{pararel,lanka,lama}, we use cloze-style questions to probe the factual knowledge in PLMs. The PLMs need to recall the captured factual knowledge to fill in the masks. Cloze-style QA uses the same input-output format as MLM. So we do not need to fine-tune the PLMs and evaluate the factual knowledge capture performance directly.

Every cloze-style question is obtained by instantiating a artificial template on a fact. For example, the question ``\textit{Keanu Reeves is a citizen of} \mask'' is constructed based on the template ``{\sc [x]} \textit{is a citizen of} {\sc [y]}'' and the triplet (\textit{Keanu Reeves}, \textit{citizen of}, \textit{Canada}). Filling the mask with the correct token ``\textit{Canada}'' is regarded as successfully capturing the corresponding fact. 

We use the cloze-style questions and evaluation metrics from \textsc{ParaRel} \cite{pararel}. \textsc{ParaRel} queries every fact with 8.64 different prompts on average. The prediction consistency is calculated by putting two different prompts into a prompt pair first (e.g., $n(n-1)/2$ pairs can be obtained from $n$ different prompts). Then the percentage of the prompt pairs that can obtain the same result is used to indicate the consistency quantitatively. The overall factual knowledge capture performance is measured by jointly considering the prediction accuracy and consistency. Table \ref{tab:cloze-data-pararel} shows the statistics of the data from \textsc{ParaRel}.

\begin{table}[htbp]
    \centering
    \small
    \begin{tabular}{lc}
    \toprule
                                & \textbf{Value} \\
    \midrule
    \# Cloze-style questions    & 199,446 \\
    \# Facts (Triplets)         & 23,097  \\
    Avg. \# questions per fact  & 8.64  \\
    
    
    \bottomrule
    \end{tabular}
    \caption{Statistics of the cloze-style QA dataset in \textsc{ParaRel} \cite{pararel}.}
    \label{tab:cloze-data-pararel}
\end{table}

\begin{table}[htbp]
    \centering
    \small
    \begin{tabular}{lc}

    \toprule
    
    
    \textbf{Dataset}      & \textbf{\# Facts } \\
    \midrule
    \textbf{LAMA}                  & 29,522  \\
    \textbf{LAMA w/o leakage}      & 25,698  \\
    \midrule
    \textbf{WIKI-UNI}              & 69,761  \\
    \textbf{WIKI-UNI w/o leakage}  & 63,772  \\
    
    
    \bottomrule
    \end{tabular}
    \caption{Statistics of the four cloze-style QA datasets from \cite{lanka}.}
    \label{tab:cloze-data-lanka}
\end{table}

We also employ four cloze-style datasets from \cite{lanka}. Table~\ref{tab:cloze-data-lanka} shows the corresponding statistics. LAMA represents the cloze-style datasets that are similar to \cite{lama} in \cite{lanka}. The distribution of the ground-truth answers in LAMA is imbalanced, providing a shortcut for PLMs to achieve good performance by selecting high-frequency entities as output. Therefore, \citet{lanka} proposes the WIKI-UNI dataset, where the distribution of the ground-truth answers follows a uniform distribution. The ground-truth answer has literal overlaps with the question sometimes. For example, in ``\textit{New York University is located in} \mask \textit{city},'' the right answer ``\textit{New York}'' exactly leaked in the question. So \citet{lanka} filter out the questions that overlap with answers and obtain two more datasets based on LAMA and WIKI-UNI, which are indicated with the suffix ``w/o leakage''.

\subsubsection{Closed-book QA}
We also use the closed-book QA to test the ability of PLMs to capture factual knowledge. As proposed in \cite{close-book-QA-BART, close-book-QA-T5,QA-overlap}, closed-book QA is similar to the way in which a student is taking a closed-book exam. The input of the model is the question only, and the model needs to generate the answer directly without seeing any other evidence. This task needs the model to generate arbitrary strings as answers, so we employ the BART, which has a decoder that can generate texts, as the base model for this task. We use the closed-book QA datasets from \cite{dpr}, as shown in Table~\ref{tab:closed-book}.

\begin{table}[htbp]
    \centering
    \small

    \begin{tabular}{lccc}
    \toprule
    \textbf{Dataset}            &  \textbf{Train Set} &   \textbf{Dev Set} &   \textbf{Test Set} \\
    \midrule
    \textbf{NaturalQuestions}   &  79,168             &  8,757             &   3,610             \\
    \textbf{TriviaQA}           &  78,785             &  8,837             &  11,313             \\
    \textbf{WebQuestions}       &   3,778             &  -\footnotemark    &   2,032             \\
    \bottomrule
    \end{tabular}

    \caption{Statistics of the closed-book QA datasets.}
    \label{tab:closed-book}
\end{table}
\footnotetext{The WebQuestions in \cite{dpr} does not have a development set, so we use the test set for both test and development.}

\subsubsection{Extractive QA}
The extractive QA is also known as machine reading comprehension \cite{MRC}. The task is to search and extract the answer span from the input passage for the input question. It evaluates the ability of PLMs to understand the facts provided in passages. We employ the six extractive QA datasets from MRQA~\cite{mrqa2019}, Table~\ref{tab:mrqa} presents the summary of the datasets. Following the setting in \cite{MRQA-script}, we use the development sets for testing.

\begin{table}[htbp]
    \centering
    \small

    \begin{tabular}{lcc}
    \toprule
    \textbf{Dataset}                  &  \textbf{Train Set} & \textbf{Test Set} \\
    \midrule
    \textbf{SQuAD}                    &  86,588             & 10,507            \\
    \textbf{NewsQA}                   &  74,160             & 4,212             \\
    \textbf{TriviaQA}                 &  61,688             & 7,785             \\
    \textbf{SearchQA}                 &  117,384            & 16,980            \\
    \textbf{HotpotQA}                 &  72,928             & 5,904             \\
    \textbf{NatrualQuestions}         &  104,071            & 12,836            \\\midrule
    \textbf{Total}                    &  620,890            & 71,060            \\
    \bottomrule
    \end{tabular}

    \caption{Summary of the extractive QA datasets.}
    \label{tab:mrqa}
\end{table}

\subsection{Results}
\subsubsection{Baselines}
We continuously pre-train RoBERTa-\textit{base} \cite{roberta} and BART-\textit{base} \cite{bart} from their official checkpoints with different masking methods:
\begin{itemize}
[itemsep=1pt,topsep=1pt,parsep=2pt]
    \item \textbf{Random token:} Mask random tokens in the tokenized text \cite{BERT}.
    \item \textbf{Whole word:} Mask random words. All the tokens in the randomly selected words are masked at once \cite{ whole-word-masking}.
    \item \textbf{Salient span:} Mask a span aligned with the subject or object, both deterministic and non-deterministic samples are included. \cite{SSM}.
    \item \textbf{Deterministic:} The proposed deterministic masking that masks a deterministic object, including only deterministic samples.
\end{itemize}
The above four models are trained with the mask-filling task only. The models pre-trained with \taska and \taskb in company with \masking are denoted as ``\textbf{+ Con \& Cls}.''
To further explore the potential of the proposed methods, we train the model on the full data with all the proposed methods, denoted as ``\textbf{+ Full data}''. We also introduce KEPLER-\textit{base} as KB-enhanced baseline for comparison.

\subsubsection{Cloze-style QA}

\begin{table*}[ht]
    \centering
    \small
    \begin{tabular}{l|ccc|ccc|ccc}
    \toprule
                                         & \multicolumn{3}{c|}{\textbf{Total}}                  & \multicolumn{3}{c|}{\textbf{Out-of-domain}}              &\multicolumn{3}{c}{\textbf{In-domain}}                \\
    \textbf{}                            & \textbf{Acc.} & \textbf{Consis.} & \textbf{\underline{Joint}} &
                                           \textbf{Acc.} & \textbf{Consis.} & \textbf{\underline{Joint}} &
                                           \textbf{Acc.} & \textbf{Consis.} & \textbf{\underline{Joint}} \\
    \midrule
    RoBERTa-\textit{base}                &           39.48 &            52.05 &\underline{16.40}&           33.09 &            55.06 &\underline{15.60}&          42.97 &           54.19 &\underline{18.55}\\
    \  Random token                      &           43.44 &            58.76 &\underline{24.72}&           35.47 &            59.57 &\underline{22.64}&          47.38 &           61.04 &\underline{27.57}\\
    \  Whole word                        &           44.04 &            58.96 &\underline{25.88}&           36.67 &            60.48 &\underline{23.01}&          47.91 &           61.57 &\underline{29.40}\\
    \  Salient span                      &           43.87 &            60.48 &\underline{26.53}&           37.48 &            61.19 &\underline{22.90}&          47.72 &           63.08 &\underline{29.93}\\
    \midrule
    
    KEPLER-\textit{base}                 &           39.63 &            50.96 &\underline{17.81}&           -     &            -     &            -     &          -     &           -     &              -     \\
    \midrule

    Deterministic                        &           45.29 &            64.65 &\underline{29.37}&           38.59 &            65.39 &\underline{25.88}&          49.13 &           66.42 &\underline{32.17}\\
    + Con \& Cls                         &           46.01 &            64.53 &\underline{29.62}&           38.05 &            65.35 &\underline{26.24}&          50.26 &           65.71 &\underline{32.32}\\
    \textbf{+ Full data}                 &  \textbf{49.40} &   \textbf{67.09} &  \textbf{\underline{33.44}} &  \textbf{40.35} &   \textbf{67.69} &   \textbf{\underline{30.30}} & \textbf{54.20} &  \textbf{69.24} &      \textbf{\underline{37.17}} \\
    
    \midrule
    \midrule
    BART-\textit{base}                   &           40.43 &            52.60 &\underline{17.83}&           34.26 &            55.88 &\underline{15.86}&          44.10 &           54.85 &\underline{20.42}\\
    \  Random token                      &           42.53 &            57.03 &\underline{23.34}&           34.98 &            59.59 &\underline{21.15}&          46.75 &           59.02 &\underline{25.78}\\
    \  Whole word                        &           41.68 &            58.96 &\underline{24.38}&           34.53 &            61.24 &\underline{21.32}&          45.11 &           60.94 &\underline{26.42}\\
    \  Salient span                      &           43.16 &            60.09 &\underline{25.30}&           35.78 &            60.68 &\underline{20.83}&          47.33 &           61.78 &\underline{27.66}\\
    \midrule

    Deterministic                        &           44.13 &            64.67 &\underline{28.77}&           36.70 &            65.15 &\underline{24.45}&          48.58 &           66.25 &\underline{31.98}\\
    + Con \& Cls                         &           46.65 &            65.64 &\underline{28.89}&           39.51 &            66.31 &\underline{25.53}&          50.72 &           67.86 &\underline{32.45}\\
    \textbf{+ Full data}                 &  \textbf{49.21} &  \textbf{ 68.41} &  \textbf{\underline{33.11}} &  \textbf{41.02} &   \textbf{69.51} &   \textbf{\underline{29.33}} & \textbf{52.95} &  \textbf{69.70} &     \textbf{\underline{35.84}} \\
    \bottomrule
    \end{tabular}
    \caption{The factual knowledge capturing performance, evaluated by the cloze-style QA dataset {\sc ParaRel}. \textbf{Acc.} is the accuracy, \textbf{Consis.} denotes the prediction consistency when changing the prompts, and \textbf{Joint} denotes the metric that jointly measures accuracy and consistency. \textbf{Out-of-domain} represent the set of questions whose triplets do not appear in the pre-training.}
    \label{tab:pararel-result}
\end{table*}

\begin{table}[ht]
    \centering
    \small

    \begin{tabular}{
    l|
    p{0.11\columnwidth}<\centering
    p{0.13\columnwidth}<\centering|
    p{0.12\columnwidth}<\centering
    p{0.13\columnwidth}<\centering
    }
    \toprule
    \multirow{2}{*}{\textbf{Dataset}}  &  \multirow{2}{*}{\textbf{LAMA}} & \textbf{w/o  }   & \textbf{WIKI-} &\textbf{w/o }    \\
                                       &                                 & \textbf{Leakage} & \textbf{UNI}   &\textbf{Leakage} \\
    \midrule
    
    RoBERTa-\textit{base}              & 19.94 &  15.10 &    10.48 &     \s7.11   \\
    \  Random token                    & 25.01 &  18.18 &    14.18 &     \s9.00   \\
    \  Whole word                      & 25.66 &  18.94 &    14.42 &     \s9.27   \\
    \  Salient span                    & 29.13 &  21.43 &    14.99 &     \s9.09   \\
    
    \midrule
    KEPLER-\textit{base}               & 15.04 &  10.66 &     8.44 &      \s5.89  \\
    \midrule

    Deterministic                      & 32.96 &  25.46 &    16.28 &      10.33   \\
    + Con \& Cls                       & 32.16 &  24.88 &    16.24 &      11.03   \\
    \textbf{+ Full data}               & \textbf{35.35} &  \textbf{28.86} &    \textbf{19.36} &      \textbf{14.23}  \\  
    \midrule\midrule
    BART-\textit{base}                 & 11.77 & \s7.08 &   \s6.03 &      \s3.47 \\
    \  Random token                    & 25.39 &  17.75 &    14.01 &      \s8.24 \\
    \  Whole word                      & 25.06 &  17.21 &    14.21 &      \s8.75 \\
    \  Salient span                    & 30.07 &  22.20 &    15.56 &      \s9.60 \\

    \midrule
    Deterministic                      & 31.26 &  23.69 &    15.81 &      10.22 \\
    + Con \& Cls                       & 32.03 &  24.57 &    15.58 &      10.23 \\
    \textbf{+ Full data}               & \textbf{35.49} &  \textbf{28.98} &    \textbf{18.64} &      \textbf{13.21} \\
    \bottomrule
    \end{tabular}

    \caption{The results on cloze-style QA datasets from \cite{lanka}. The performance is measured by accuracy\footnotemark.}
    \label{tab:lanka-result}
\end{table}

\noindent\textbf{Masking strategies}
Tables~\ref{tab:pararel-result} and \ref{tab:lanka-result} present the results on cloze-style QA. We can see that random token masking can gain some improvements in performance, as well as the whole word masking. We think this is because the input texts, which come from Wikipedia, are formal descriptions of facts. Training on such texts helps shift the domain of PLMs for better generating factual words. The random token and whole word masking serve as solid baselines to focus the comparison between masking strategies, eliminating the confounders brought by extra pre-training on Wikipedia.

The salient span masking and \masking both mask entity spans. The difference is that \masking further limits the relationship between the remaining context and the masked span to be deterministic, driving PLMs to learn to infer based on the deterministic clues. The results show that the PLMs can achieve much better results with \masking, indicating that the deterministic relationship is valuable for recovering factual spans robustly. 

\noindent\textbf{The proposed pre-training tasks}
The \taska and the \taskb, which aim to strengthen the deterministic relationship, also provide further performance improvements  (denoted as \textbf{+ Con \& Cls}). Finally, the full data with all the proposed methods brings the most significant improvement. The proposed pre-training models also outperform the KEPLER-\textit{base}.

\noindent\textbf{Out-of-domain evaluation}
To analyze the improvement in-depth, we split the probing questions into \textit{in-domain} and \textit{out-of-domain} according to whether the pre-training corpus covers the corresponding triplets in questions. As Table~\ref{tab:pararel-result} shows, the three random-based masking methods (Random token, Whole word, and Salient Span) boost performance on in-domain questions but get stuck on the out-of-domain questions. It is natural that the PLMs can answer the questions that are involved in pre-training. Surprisingly, although the out-of-domain questions are inaccessible in the pre-training corpus, the \masking also gains performance improvement (3-4\%), indicating that the deterministic relationship could help PLMs to better recollect the facts learned implicitly.

\subsubsection{Closed-book QA}
We fine-tune the continuously pre-trained BART-\textit{base} on the Closed-book QA task.
The metrics are EM(Exact Match) and F1 from~\cite{squadv1}. Table~\ref{tab:closed-book-qa} shows the comparison results of different strategies.

\footnotetext{The detailed metrics grouped by the relation types (N-1,N-M relations) are in Appendix~\ref{sec:app_detail_cloze_results}.}

\begin{table}[ht]
    \centering
    \small
    \begin{tabular}{llcc}
    \toprule
    \textbf{Dataset}                      & \textbf{Model}             & \textbf{F1} & \textbf{EM} \\
    \midrule                                                                                    
    \multirow{7}{*}{TriviaQA}             & BART-\textit{base}         & 23.91       & 17.52 \\ 
                                          & \  Random token            & 23.67       & 18.04 \\ 
                                          & \  Whole word              & 24.64       & 18.88 \\ 
                                          & \  Salient span            & 24.98       & 19.21 \\ 
                                          \cmidrule{2-4}
                                          & Deterministic              & 24.94       & 19.22 \\ 
                                          & + Con \& Cls               & 25.28       & 19.58 \\ 

                                          & \textbf{+ Full Data}       & \textbf{26.35}       & \textbf{20.57} \\\midrule 
                                          
    \multirow{7}{*}{NaturalQuestions}     & BART-\textit{base}         & 26.89       & 21.27 \\ 
                                          & \  Random token            & 27.34       & 21.55 \\ 
                                          & \  Whole word              & 27.53       & 22.13 \\ 
                                          & \  Salient span            & 27.31       & 22.07 \\ 
                                          \cmidrule{2-4}
                                          & Deterministic              & 27.83       & 22.60 \\ 
                                          & + Con \& Cls               & 28.14       & 22.69 \\ 

                                          & \textbf{+ Full Data}       & \textbf{29.17}       & \textbf{23.91} \\ \midrule
                                        
    \multirow{7}{*}{WebQuestions}         & BART-\textit{base}         & 33.62       & 26.62 \\ 
                                          & \  Random token            & 32.58       & 26.38 \\ 
                                          & \  Whole word              & 32.45       & 26.03 \\ 
                                          &\   Salient span            & 32.70       & 26.08 \\ 
                                          \cmidrule{2-4}
                                          & Deterministic              & 32.73       & 26.38 \\ 
                                          & + Con \& Cls               & 32.63       & 25.59 \\ 
                                          & \textbf{+ Full Data}       & \textbf{33.91}       & \textbf{27.26} \\ 
    \bottomrule
    \end{tabular}
    \caption{The performance on the closed-book QA datasets.}
    \label{tab:closed-book-qa}
\end{table}

Closed-book QA is more difficult than cloze-style QA since the models need to generate answers without any extra hints, e.g., the answer length is indicated by the number of \mask{}s in the cloze-style QA, while the models need to predict the answer length in closed-book QA. The input-output format of closed-book QA differs from per-training, so we need to fine-tune PLMs to recall facts based on natural questions to fit this format. Table~\ref{tab:closed-book-qa} shows the evaluation results. Generally, the proposed methods outperform the baselines, demonstrating that the proposed methods can help the PLM that uses encoder-decoder architecture to capture and recall factual knowledge.

\subsubsection{Extractive QA}
We fine-tune the models that based on RoBERTa-\textit{base} for extractive QA. Following \cite{kgplm,spanbert}, we employ the MRQA data with two different settings: (a) \textbf{Separate:} the models are trained and tested on every QA dataset separately, (b) \textbf{Combine:} all the training samples from the six datasets are merged in training. Then the fine-tuned models are evaluated on each dataset respectively. Table~\ref{tab:extractive-qa} shows the evaluation results, the metrics are averaged over the six development sets.

\begin{table}[htbp]
    \centering
    \small
    \begin{tabular}{l|cc|cc}
    \toprule
    \multirow{2}{*}{\textbf{Model}}           & \multicolumn{2}{c|}{\textbf{Separate}}       & \multicolumn{2}{c}{\textbf{Combine}} \\
    {}                                        & \textbf{F1} & \textbf{EM}                    & \textbf{F1}     & \textbf{EM}        \\
    \midrule
    RoBERTa-\textit{base}                     & 80.78       & 69.51                          &           81.78 &           70.57    \\

    \  Random token                           & 80.53       & 69.22                          &           81.79 &           70.52    \\
    \  Whole word                             & 80.86       & 69.71                          &           81.77 &           70.60    \\
    \  Salient span                           & 80.85       & 69.61                          &           81.72 &           70.50    \\
    \midrule
    KEPLER-\textit{base}                      & 80.28       & 69.02                          &           81.41 &           70.32    \\

    \midrule
    Deterministic                             & 80.83       & 69.63                          &           81.78 &           70.59    \\
    
    + Con \& Cls                              & 80.94       & 69.75                          &           81.79 &           70.67    \\ 
    \textbf{+ Full data}                      & \textbf{80.96} & \textbf{69.67}              &  \textbf{81.86} & \textbf{ 70.71}    \\

    \bottomrule
    \end{tabular}
    \caption{The performance on the extractive QA task.}
    \label{tab:extractive-qa}
\end{table}

In extractive QA, the input includes a question and the supporting evidence to answer it. So the models do not have to recollect the essential evidence but should put more effort into understanding the evidence. Table~\ref{tab:extractive-qa} shows the evaluation results. Due to the difference in the input-output format between the MLM and span extraction task, the change in the masking methods has somewhat limited effects on the performance here. The averaging on six different MRC datasets and the hyperparameter search (in Appendix~\ref{sec:app_hp}) could further diminish the performance difference between the models. However, the proposed methods still show slight advantages in the comparison, demonstrating that learning the deterministic relationship could also help to comprehend factual knowledge.

\section{Related Work}
Pre-training on large-scale unlabeled text can help PLMs capture meaningful knowledge and benefits the downstream tasks accordingly \cite{GPT,radford2018improving}. BERT \cite{BERT} proposes a Mask Language Model (MLM) in which the model needs to recover some masked tokens based on the remaining context. The effectiveness of the MLM makes BERT become the starting point for fitting many downstream tasks \cite{chen2020lottery}.
Afterward, several different masking methods \cite{whole-word-masking,spanbert,pmi-masking,ERNIE} have explored how masking methods affect performance and have obtained further performance improvement. These works push the limit of MLM and show the importance of designing better masking strategies.

On the other hand, some pre-training tasks other than MLM have been proposed. \citet{electra} trains the model to distinguish the replaced words from the original words in the context. \cite{WKLM,kgplm} let factual spans be the replacement candidates. \citet{ERICA} contrasts the representations between different entities and relations. This paper views another perspective of the masking methods: whether the remaining context can uniquely determine the masked span. Accordingly, we propose a deterministic masking strategy that masks deterministic spans in MLM samples. Moreover, we design \taska and \taskb as pre-training tasks to help PLMs identify the deterministic clues for the masked span and contrast them with the non-deterministic ones. Moreover, we evaluate the performance of the proposed model with various downstream tasks.

\section{Conclusion}

This paper proposes to train PLMs to learn a deterministic input-output relationship in MLM to improve PLMs on capturing factual knowledge. The deterministic relationship ensures the masked content in MLM samples is deterministically predictable based on the remaining context. To further enhance the deterministic relationship, we design a pre-training task \taska that encourages PLMs to give more confident predictions when the input keeps deterministic clues, and the \taskb to train PLMs to predict whether the deterministic clues exist. Extensive experiments show that the proposed methods can improve the accuracy and consistency of factual knowledge capturing and boost the performance of the other two knowledge-intensive tasks.

\section{Limitations}
We summarize this paper's main limitations as follows:
First, this study focuses on enhancing the deterministic relationship but does not explore the non-deterministic relationships. The other non-deterministic relationships also play essential roles in tasks such as semantic role labeling and emotion recognition, where the proposed methods may not be helpful. Second, due to the diversity and richness of natural language, we cannot perfectly recognize the deterministic clues and spans from texts. We have to consider the noises in recognization when designing the pre-training tasks. Finally, we continuously pre-train PLMs on only Wikipedia text, somewhat narrowing down their domain. Constructing more pre-training samples by the proposed procedure (Procedure \ref{proc:data_construction} in the Appendix) could be better. Moreover, we can use the current pre-training samples (based on Wikipedia) to train an ``interpolation model'' that can tag the deterministic clues and spans in the input texts. The interpolation model can also be used to enlarge the pre-training data.

\section*{Acknowledgements}
We would like to thank the anonymous reviewers for providing valuable reviews throughout the multi-turn rolling review progress. Thanks to Benyou Wang for the helpful discussions, suggestions, and encouragement.

\bibliography{anthology,custom}

\begin{thebibliography}{41}
\expandafter\ifx\csname natexlab\endcsname\relax\def\natexlab#1{#1}\fi

\bibitem[{Brown et~al.(2020)Brown, Mann, Ryder, Subbiah, Kaplan, Dhariwal,
  Neelakantan, Shyam, Sastry, Askell et~al.}]{GPT}
Tom~B Brown, Benjamin Mann, Nick Ryder, Melanie Subbiah, Jared Kaplan, Prafulla
  Dhariwal, Arvind Neelakantan, Pranav Shyam, Girish Sastry, Amanda Askell,
  et~al. 2020.
\newblock Language models are few-shot learners.
\newblock \emph{arXiv preprint arXiv:2005.14165}.

\bibitem[{Cao et~al.(2021)Cao, Lin, Han, Sun, Yan, Liao, Xue, and Xu}]{lanka}
Boxi Cao, Hongyu Lin, Xianpei Han, Le~Sun, Lingyong Yan, Meng Liao, Tong Xue,
  and Jin Xu. 2021.
\newblock \href {https://aclanthology.org/2021.acl-long.146} {Knowledgeable or
  educated guess? revisiting language models as knowledge bases}.
\newblock In \emph{Proceedings of the 59th Annual Meeting of the Association
  for Computational Linguistics and the 11th International Joint Conference on
  Natural Language Processing (Volume 1: Long Papers)}, pages 1860--1874,
  Online. Association for Computational Linguistics.

\bibitem[{Chen et~al.(2020)Chen, Frankle, Chang, Liu, Zhang, Wang, and
  Carbin}]{chen2020lottery}
Tianlong Chen, Jonathan Frankle, Shiyu Chang, Sijia Liu, Yang Zhang, Zhangyang
  Wang, and Michael Carbin. 2020.
\newblock The lottery ticket hypothesis for pre-trained bert networks.
\newblock \emph{arXiv preprint arXiv:2007.12223}.

\bibitem[{Clark et~al.(2019)Clark, Luong, Le, and Manning}]{electra}
Kevin Clark, Minh-Thang Luong, Quoc~V Le, and Christopher~D Manning. 2019.
\newblock Electra: Pre-training text encoders as discriminators rather than
  generators.
\newblock In \emph{International Conference on Learning Representations}.

\bibitem[{Cole et~al.(2021)Cole, Mac~Aodha, Lorieul, Perona, Morris, and
  Jojic}]{multi-label-false-negative}
Elijah Cole, Oisin Mac~Aodha, Titouan Lorieul, Pietro Perona, Dan Morris, and
  Nebojsa Jojic. 2021.
\newblock Multi-label learning from single positive labels.
\newblock In \emph{Proceedings of the IEEE/CVF Conference on Computer Vision
  and Pattern Recognition}, pages 933--942.

\bibitem[{Cover(1999)}]{information-theory}
Thomas~M Cover. 1999.
\newblock \emph{Elements of information theory}.
\newblock John Wiley \& Sons.

\bibitem[{Cui et~al.(2019)Cui, Che, Liu, Qin, Yang, Wang, and
  Hu}]{whole-word-masking}
Yiming Cui, Wanxiang Che, Ting Liu, Bing Qin, Ziqing Yang, Shijin Wang, and
  Guoping Hu. 2019.
\newblock Pre-training with whole word masking for chinese bert.
\newblock \emph{arXiv preprint arXiv:1906.08101}.

\bibitem[{Devlin et~al.(2018)Devlin, Chang, Lee, and Toutanova}]{BERT}
Jacob Devlin, Ming{-}Wei Chang, Kenton Lee, and Kristina Toutanova. 2018.
\newblock \href {http://arxiv.org/abs/1810.04805} {{BERT:} pre-training of deep
  bidirectional transformers for language understanding}.
\newblock \emph{CoRR}, abs/1810.04805.

\bibitem[{Durand et~al.(2019)Durand, Mehrasa, and
  Mori}]{multi-label-missing-multi-labels}
Thibaut Durand, Nazanin Mehrasa, and Greg Mori. 2019.
\newblock Learning a deep convnet for multi-label classification with partial
  labels.
\newblock In \emph{Proceedings of the IEEE/CVF Conference on Computer Vision
  and Pattern Recognition}, pages 647--657.

\bibitem[{Elazar et~al.(2021)Elazar, Kassner, Ravfogel, Ravichander, Hovy,
  Schütze, and Goldberg}]{pararel}
Yanai Elazar, Nora Kassner, Shauli Ravfogel, Abhilasha Ravichander, Eduard
  Hovy, Hinrich Schütze, and Yoav Goldberg. 2021.
\newblock \href {http://arxiv.org/abs/2102.01017} {Measuring and improving
  consistency in pretrained language models}.

\bibitem[{Elsahar et~al.(2018)Elsahar, Vougiouklis, Remaci, Gravier, Hare,
  Laforest, and Simperl}]{TREX}
Hady Elsahar, Pavlos Vougiouklis, Arslen Remaci, Christophe Gravier, Jonathon
  Hare, Frederique Laforest, and Elena Simperl. 2018.
\newblock T-rex: A large scale alignment of natural language with knowledge
  base triples.
\newblock In \emph{Proceedings of the Eleventh International Conference on
  Language Resources and Evaluation (LREC 2018)}.

\bibitem[{Fisch et~al.(2019)Fisch, Talmor, Jia, Seo, Choi, and Chen}]{mrqa2019}
Adam Fisch, Alon Talmor, Robin Jia, Minjoon Seo, Eunsol Choi, and Danqi Chen.
  2019.
\newblock {MRQA} 2019 shared task: Evaluating generalization in reading
  comprehension.
\newblock In \emph{Proceedings of 2nd Machine Reading for Reading Comprehension
  (MRQA) Workshop at EMNLP}.

\bibitem[{Guu et~al.(2020)Guu, Lee, Tung, Pasupat, and Chang}]{SSM}
Kelvin Guu, Kenton Lee, Zora Tung, Panupong Pasupat, and Ming-Wei Chang. 2020.
\newblock Realm: Retrieval-augmented language model pre-training.
\newblock \emph{arXiv preprint arXiv:2002.08909}.

\bibitem[{He et~al.(2020)He, Jiang, Xiao, and Liu}]{kgplm}
Bin He, Xin Jiang, Jinghui Xiao, and Qun Liu. 2020.
\newblock Kgplm: Knowledge-guided language model pre-training via generative
  and discriminative learning.
\newblock \emph{arXiv preprint arXiv:2012.03551}.

\bibitem[{Jiang et~al.(2020)Jiang, Xu, Araki, and Neubig}]{JiangXAN20}
Zhengbao Jiang, Frank~F. Xu, Jun Araki, and Graham Neubig. 2020.
\newblock How can we know what language models know.
\newblock \emph{Trans. Assoc. Comput. Linguistics}, 8:423--438.

\bibitem[{Joshi et~al.(2020)Joshi, Chen, Liu, Weld, Zettlemoyer, and
  Levy}]{spanbert}
Mandar Joshi, Danqi Chen, Yinhan Liu, Daniel~S Weld, Luke Zettlemoyer, and Omer
  Levy. 2020.
\newblock Spanbert: Improving pre-training by representing and predicting
  spans.
\newblock \emph{Transactions of the Association for Computational Linguistics},
  8:64--77.

\bibitem[{Karpukhin et~al.(2020)Karpukhin, Oguz, Min, Lewis, Wu, Edunov, Chen,
  and Yih}]{dpr}
Vladimir Karpukhin, Barlas Oguz, Sewon Min, Patrick Lewis, Ledell Wu, Sergey
  Edunov, Danqi Chen, and Wen-tau Yih. 2020.
\newblock Dense passage retrieval for open-domain question answering.
\newblock In \emph{Proceedings of the 2020 Conference on Empirical Methods in
  Natural Language Processing (EMNLP)}, pages 6769--6781.

\bibitem[{Levine et~al.(2020)Levine, Lenz, Lieber, Abend, Leyton-Brown,
  Tennenholtz, and Shoham}]{pmi-masking}
Yoav Levine, Barak Lenz, Opher Lieber, Omri Abend, Kevin Leyton-Brown, Moshe
  Tennenholtz, and Yoav Shoham. 2020.
\newblock Pmi-masking: Principled masking of correlated spans.
\newblock In \emph{International Conference on Learning Representations}.

\bibitem[{Lewis et~al.(2020)Lewis, Liu, Goyal, Ghazvininejad, Mohamed, Levy,
  Stoyanov, and Zettlemoyer}]{bart}
Mike Lewis, Yinhan Liu, Naman Goyal, Marjan Ghazvininejad, Abdelrahman Mohamed,
  Omer Levy, Veselin Stoyanov, and Luke Zettlemoyer. 2020.
\newblock Bart: Denoising sequence-to-sequence pre-training for natural
  language generation, translation, and comprehension.
\newblock In \emph{Proceedings of the 58th Annual Meeting of the Association
  for Computational Linguistics}, pages 7871--7880.

\bibitem[{Lewis et~al.(2021)Lewis, Stenetorp, and Riedel}]{QA-overlap}
Patrick Lewis, Pontus Stenetorp, and Sebastian Riedel. 2021.
\newblock Question and answer test-train overlap in open-domain question
  answering datasets.
\newblock In \emph{Proceedings of the 16th Conference of the European Chapter
  of the Association for Computational Linguistics: Main Volume}, pages
  1000--1008.

\bibitem[{Liu et~al.(2019{\natexlab{a}})Liu, Zhang, Zhang, Wang, and
  Zhang}]{MRC}
Shanshan Liu, Xin Zhang, Sheng Zhang, Hui Wang, and Weiming Zhang.
  2019{\natexlab{a}}.
\newblock Neural machine reading comprehension: Methods and trends.
\newblock \emph{Applied Sciences}, 9(18):3698.

\bibitem[{Liu et~al.(2019{\natexlab{b}})Liu, Ott, Goyal, Du, Joshi, Chen, Levy,
  Lewis, Zettlemoyer, and Stoyanov}]{roberta}
Yinhan Liu, Myle Ott, Naman Goyal, Jingfei Du, Mandar Joshi, Danqi Chen, Omer
  Levy, Mike Lewis, Luke Zettlemoyer, and Veselin Stoyanov. 2019{\natexlab{b}}.
\newblock Roberta: A robustly optimized bert pretraining approach.
\newblock \emph{arXiv preprint arXiv:1907.11692}.

\bibitem[{Luo et~al.(2021)Luo, Chen, Tan, Li, He, and
  Jia}]{entropy-as-certainty-2}
Xin Luo, Wei Chen, Yusong Tan, Chen Li, Yulin He, and Xiaogang Jia. 2021.
\newblock Exploiting negative learning for implicit pseudo label rectification
  in source-free domain adaptive semantic segmentation.
\newblock \emph{arXiv preprint arXiv:2106.12123}.

\bibitem[{Petroni et~al.(2019)Petroni, Rockt{\"{a}}schel, Riedel, Lewis,
  Bakhtin, Wu, and Miller}]{lama}
Fabio Petroni, Tim Rockt{\"{a}}schel, Sebastian Riedel, Patrick S.~H. Lewis,
  Anton Bakhtin, Yuxiang Wu, and Alexander~H. Miller. 2019.
\newblock Language models as knowledge bases?
\newblock In \emph{Proceedings of the 2019 Conference on Empirical Methods in
  Natural Language Processing and the 9th International Joint Conference on
  Natural Language Processing, {EMNLP-IJCNLP} 2019, Hong Kong, China, November
  3-7, 2019}, pages 2463--2473.

\bibitem[{P{\"{o}}rner et~al.(2020)P{\"{o}}rner, Waltinger, and
  Sch{\"{u}}tze}]{DBLP:e-BERT}
Nina P{\"{o}}rner, Ulli Waltinger, and Hinrich Sch{\"{u}}tze. 2020.
\newblock {E-BERT:} efficient-yet-effective entity embeddings for {BERT}.
\newblock In \emph{Findings of the Association for Computational Linguistics:
  {EMNLP} 2020, Online Event, 16-20 November 2020}, volume {EMNLP} 2020.

\bibitem[{Qin et~al.(2020)Qin, Lin, Takanobu, Liu, Li, Ji, Huang, Sun, and
  Zhou}]{ERICA}
Yujia Qin, Yankai Lin, Ryuichi Takanobu, Zhiyuan Liu, Peng Li, Heng Ji, Minlie
  Huang, Maosong Sun, and Jie Zhou. 2020.
\newblock Erica: improving entity and relation understanding for pre-trained
  language models via contrastive learning.
\newblock \emph{arXiv preprint arXiv:2012.15022}.

\bibitem[{Radford et~al.(2018)Radford, Narasimhan, Salimans, and
  Sutskever}]{radford2018improving}
Alec Radford, Karthik Narasimhan, Tim Salimans, and Ilya Sutskever. 2018.
\newblock Improving language understanding by generative pre-training.

\bibitem[{Rajpurkar et~al.(2016)Rajpurkar, Zhang, Lopyrev, and Liang}]{squadv1}
Pranav Rajpurkar, Jian Zhang, Konstantin Lopyrev, and Percy Liang. 2016.
\newblock Squad: 100,000+ questions for machine comprehension of text.
\newblock In \emph{Proceedings of the 2016 Conference on Empirical Methods in
  Natural Language Processing}, pages 2383--2392.

\bibitem[{Ram et~al.(2021{\natexlab{a}})Ram, Kirstain, Berant, Globerson, and
  Levy}]{MRQA-script}
Ori Ram, Yuval Kirstain, Jonathan Berant, Amir Globerson, and Omer Levy.
  2021{\natexlab{a}}.
\newblock Few-shot question answering by pretraining span selection.
\newblock \emph{arXiv preprint arXiv:2101.00438}.

\bibitem[{Ram et~al.(2021{\natexlab{b}})Ram, Kirstain, Berant, Globerson, and
  Levy}]{mrqa-code-source}
Ori Ram, Yuval Kirstain, Jonathan Berant, Amir Globerson, and Omer Levy.
  2021{\natexlab{b}}.
\newblock \href {https://aclanthology.org/2021.acl-long.239} {Few-shot question
  answering by pretraining span selection}.
\newblock In \emph{Proceedings of the 59th Annual Meeting of the Association
  for Computational Linguistics and the 11th International Joint Conference on
  Natural Language Processing (Volume 1: Long Papers)}, pages 3066--3079,
  Online. Association for Computational Linguistics.

\bibitem[{Reiter(1981)}]{closed-world-assumption}
Raymond Reiter. 1981.
\newblock On closed world data bases.
\newblock In \emph{Readings in artificial intelligence}, pages 119--140.
  Elsevier.

\bibitem[{Roberts et~al.(2020)Roberts, Raffel, and Shazeer}]{close-book-QA-T5}
Adam Roberts, Colin Raffel, and Noam Shazeer. 2020.
\newblock How much knowledge can you pack into the parameters of a language
  model?
\newblock In \emph{Proceedings of the 2020 Conference on Empirical Methods in
  Natural Language Processing (EMNLP)}, pages 5418--5426.

\bibitem[{Shin et~al.(2020)Shin, Razeghi, Logan~IV, Wallace, and
  Singh}]{shin2020autoprompt}
Taylor Shin, Yasaman Razeghi, Robert~L Logan~IV, Eric Wallace, and Sameer
  Singh. 2020.
\newblock Autoprompt: Automatic prompt construction for masked language models.
\newblock In \emph{Empirical Methods in Natural Language Processing (EMNLP)},
  pages 4222--4235.

\bibitem[{Sun et~al.(2019)Sun, Wang, Li, Feng, Chen, Zhang, Tian, Zhu, Tian,
  and Wu}]{ERNIE}
Yu~Sun, Shuohuan Wang, Yukun Li, Shikun Feng, Xuyi Chen, Han Zhang, Xin Tian,
  Danxiang Zhu, Hao Tian, and Hua Wu. 2019.
\newblock Ernie: Enhanced representation through knowledge integration.
\newblock \emph{arXiv preprint arXiv:1904.09223}.

\bibitem[{van Hulst et~al.(2020)van Hulst, Hasibi, Dercksen, Balog, and
  de~Vries}]{entity-linker-REL}
Johannes~M. van Hulst, Faegheh Hasibi, Koen Dercksen, Krisztian Balog, and
  Arjen~P. de~Vries. 2020.
\newblock Rel: An entity linker standing on the shoulders of giants.
\newblock In \emph{Proceedings of the 43rd International ACM SIGIR Conference
  on Research and Development in Information Retrieval}, SIGIR '20. ACM.

\bibitem[{Vaswani et~al.(2017)Vaswani, Shazeer, Parmar, Uszkoreit, Jones,
  Gomez, Kaiser, and Polosukhin}]{transformer}
Ashish Vaswani, Noam Shazeer, Niki Parmar, Jakob Uszkoreit, Llion Jones,
  Aidan~N Gomez, {\L}ukasz Kaiser, and Illia Polosukhin. 2017.
\newblock Attention is all you need.
\newblock In \emph{Advances in neural information processing systems}, pages
  5998--6008.

\bibitem[{Vu et~al.(2019)Vu, Jain, Bucher, Cord, and
  P{\'e}rez}]{entropy-as-certainty-1}
Tuan-Hung Vu, Himalaya Jain, Maxime Bucher, Matthieu Cord, and Patrick
  P{\'e}rez. 2019.
\newblock Advent: Adversarial entropy minimization for domain adaptation in
  semantic segmentation.
\newblock In \emph{Proceedings of the IEEE/CVF Conference on Computer Vision
  and Pattern Recognition}, pages 2517--2526.

\bibitem[{Wang et~al.(2021)Wang, Liu, and Zhang}]{close-book-QA-BART}
Cunxiang Wang, Pai Liu, and Yue Zhang. 2021.
\newblock Can generative pre-trained language models serve as knowledge bases
  for closed-book qa?
\newblock \emph{arXiv preprint arXiv:2106.01561}.

\bibitem[{Xiong et~al.(2019)Xiong, Du, Wang, and Stoyanov}]{WKLM}
Wenhan Xiong, Jingfei Du, William~Yang Wang, and Veselin Stoyanov. 2019.
\newblock Pretrained encyclopedia: Weakly supervised knowledge-pretrained
  language model.
\newblock \emph{arXiv preprint arXiv:1912.09637}.

\bibitem[{Zhang and Zhou(2006)}]{multi-label-def}
Min-Ling Zhang and Zhi-Hua Zhou. 2006.
\newblock Multilabel neural networks with applications to functional genomics
  and text categorization.
\newblock \emph{IEEE transactions on Knowledge and Data Engineering},
  18(10):1338--1351.

\bibitem[{Zhong et~al.(2021)Zhong, Friedman, and Chen}]{optim-prompt}
Zexuan Zhong, Dan Friedman, and Danqi Chen. 2021.
\newblock Factual probing is {[MASK]:} learning vs. learning to recall.
\newblock \emph{CoRR}, abs/2104.05240.

\end{thebibliography}
\bibliographystyle{acl_natbib}
\clearpage

\appendix
\section*{Appendix}
\section{Ablation Study}
\label{sec:app_detail_cloze_results}

\noindent\textbf{How does masking objects help in factual knowledge capturing?}
As described in Section~\ref{sec:task-cloze-style-qa}, the cloze-style questions that we use to probe the factual knowledge in PLMs, are constructed by integrating subject-predicate-object triples with artificial templates. 
Due to the conventions in the template construction, the objects could have more opportunities to be the answer than the predicates and subjects. So focusing on recovering objects in pre-training may also benefit cloze-style QA. The proposed \masking naturally masks more objects in pre-training because of the rules we designed to identify the deterministic span. Both masking objects and the deterministic relationship could bring improvements in the \masking.

To investigate and clarify their contributions, we introduce a baseline ``Object'' that masks and predicts only object spans in pre-training. Table~\ref{tab:pararel-objects-cls-con-result} shows the evaluation result. We can see that the Object baseline performs better than the Salient span baseline on factual knowledge capture. It reveals that masking objects indeed improve performance. Nevertheless, the \masking (denoted as ``Deterministic'') achieves better results, denoting that both masking objects and learning the deterministic relationship contribute positively to factual knowledge capture.

\begin{table*}[ht]
    \centering
    \small
    \begin{tabular}{ll|ccc|ccc|ccc}
    \toprule
    \multicolumn{2}{c|}{}       &  \multicolumn{3}{c|}{\textbf{Total}}                 & \multicolumn{3}{c|}{\textbf{Out-of-domain}}           &\multicolumn{3}{c}{\textbf{In-domain}}             \\
    \multicolumn{2}{c|}{}       &  \textbf{Acc.} & \textbf{Consis.} & \textbf{\underline{Joint}} &
                                   \textbf{Acc.} & \textbf{Consis.} & \textbf{\underline{Joint}} &
                                   \textbf{Acc.} & \textbf{Consis.} & \textbf{\underline{Joint}} \\
    \midrule
    \multirow{5}{*}{\textbf{RoBERTa-\textit{base}}}
    & Salient span              &           43.87 &            60.48 &\underline{26.53}&           37.48 &            61.19 &\underline{22.90}&          47.72 &           63.08 &\underline{29.93} \\
    & Object                    &           44.41 &            63.68 &\underline{28.00}&           38.48 &            64.40 &\underline{24.94}&          47.82 &           65.25 &\underline{31.27} \\
    & Deterministic             &           45.29 &            64.65 &\underline{29.37}&           38.59 &            65.39 &\underline{25.88}&          49.13 &           66.42 &\underline{32.17} \\
    & + Con                     &           45.93 &            65.50 &\underline{29.52}&           39.33 &            66.34 &\underline{26.26}&          49.50 &           67.26 &\underline{32.58} \\ 
    & + Con \& Cls              &           46.01 &            64.53 &\underline{29.62}&           38.05 &            65.35 &\underline{26.24}&          50.26 &           65.71 &\underline{32.32}\\

    \midrule
    \midrule
    \multirow{5}{*}{\textbf{BART-\textit{base}}}
    & Salient span              &           43.16 &            60.09 &\underline{25.30}&           35.78 &            60.68 &\underline{20.83}&          47.33 &           61.78 &\underline{27.66}\\
    & Object                    &           42.61 &            62.67 &\underline{27.24}&           35.76 &            62.22 &\underline{22.39}&          46.91 &           64.89 &\underline{30.73}\\
    & Deterministic             &           44.13 &            64.67 &\underline{28.77}&           36.70 &            65.15 &\underline{24.45}&          48.58 &           66.25 &\underline{31.98}\\
    & + Con                     &           44.96 &            64.11 &\underline{28.17}&           37.66 &            64.50 &\underline{24.51}&          48.97 &           66.16 &\underline{31.71}\\
    & + Con \& Cls              &           46.65 &            65.64 &\underline{28.89}&           39.51 &            66.31 &\underline{25.53}&          50.72 &           67.86 &\underline{32.45}\\
    \bottomrule
    \end{tabular}
    \caption{The evaluation results on {\sc ParaRel}. ``Object'' denotes the baseline that masks and predicts the objects. ``Deterministic'' denotes the MLM baseline that uses deterministic masking. ``+ Con'' is the baseline that uses the \taska with \masking.}
    \label{tab:pararel-objects-cls-con-result}
    
\end{table*}

\noindent\textbf{The effectiveness of the \taska and \taskb}
To reveal the contribution of the proposed pre-training tasks separately, we introduce a baseline that only uses the \taska. We refer to it as ``+ Con'' in Table~\ref{tab:pararel-objects-cls-con-result}. ``+ Con \& Cls'' denotes the PLM that uses the \taskb in addition to the \taska. We can see that the performance increases as we apply the proposed methods incrementally.

\noindent \textbf{The improvements on deterministic and non-deterministic cloze-style questions}
The dataset from \textsc{ParaRel} includes only the N-1 relations\footnotemark.
\footnotetext{Defined in {https://github.com/yanaiela/pararel/wiki/31-N1-Relations}}
While the LAMA dataset from \cite{lama} (Tables \ref{tab:cloze-data-lanka} and \ref{tab:lanka-result}) includes both the N-1/1-1 and N-M relations. To reveal the improvements in terms of relation types, we separate the samples into N-1/1-1 and N-M based on the relation types and report the results separately, as shown in Table~\ref{tab:detail-lama-result}. Similar to the deterministic relationship we used, the ground-truth object is unique when the relation type is N-1 or 1-1. The results show that the improvement for the N-1/1-1 relations is more significant than the N-M relations when using the proposed methods.

\begin{table}[htbp]
    \centering
    \small

    \begin{tabular}{l|ccc
    }
    \toprule
    \textbf{Model}                     & \textbf{Total}  & \textbf{N-1/1-1 } & \textbf{N-M}    \\
    \midrule
    
    RoBERTa-\textit{base}              & 19.94 &      22.18 &          16.46 \\
    \  Random token                    & 25.01 &      28.98 &          18.81 \\
    \  Whole word                      & 25.66 &      29.26 &          20.05 \\
    \  Salient span                    & 29.13 &      30.89 &          26.38 \\
    \midrule
    
    Deterministic                      & 32.96 &      37.84 &          25.35 \\
    + Con \& Cls                       & 32.16 &      36.62 &          25.19 \\
    
    \textbf{+ Full data}               & \textbf{35.35} &      \textbf{41.86} &          \textbf{25.18}  \\  
    \midrule\midrule
    
    BART-\textit{base}                 & 11.77 &      13.96 &          \s8.35 \\
    \  Random token                    & 25.39 &      27.66 &          21.84  \\
    \  Whole word                      & 25.06 &      28.69 &          19.39  \\
    \  Salient span                    & 30.07 &      31.83 &          27.31  \\
    \midrule
    Deterministic                      & 31.26 &      35.77 &          24.21  \\
    + Con \& Cls                       & 32.03 &      36.38 &          25.25  \\
    \textbf{+ Full data}               & \textbf{35.49} &      \textbf{41.99} &          \textbf{25.33}  \\

    \bottomrule
    \end{tabular}

    \caption{The detailed results on the LAMA dataset in \cite{lanka}, reported separately with respect to relation types: N-1/1-1 or N-M.}
    \label{tab:detail-lama-result}
\end{table}

\section{Pre-training Data Construction}
\label{sec:app_pt_data_cons}

\begin{algorithm*}[htbp]
    \floatname{algorithm}{Procedure}
    \renewcommand{\algorithmicensure}{\textbf{Output:}}
    \renewcommand{\algorithmicprocedure}{\textbf{function}}
    \caption{Deterministic Sample Construction}
    \label{proc:data_construction}
    \begin{algorithmic}[1]
        \Require {
        Text collection $T$, 
        Knowledge base $K$, 
        ${\textsc{EntityLinker}}$
        }
        \Ensure Deterministic sample collection $D_{\rm d}$
        \Ensure Salient span masking sample collection $D_{\rm ssm}$
        \State $D_{\rm d} \gets \{\}$,  $D_{\rm ssm} \gets \{\}$
        \ForAll {text piece $t$ \textbf{in} $T$}
            \State $E \gets {\textsc{EntityLinker}}(t)$                       \Comment{Identify all the entities in $t$}
            \State $D_{\rm ssm} = D_{\rm ssm} \cup \{(t, E)\}$                  \Comment{Save the entities for the salient span masking}
            \ForAll {entity pair $(e_i,e_j)$ \textbf{in} $E\times E$}
                \ForAll {predicate $r$ that can connects $(e_i,e_j)$}         \Comment{Triplet $(e_i,r,e_j)$ exists in $K$}
                \If {$e_j$ has only one match when querying $K$ with $e_i$ and $r$ as the subject and predicate} 
                    \State $p \gets \textsc{PredicateMatcher}(t,r)$           \Comment{Find the spans in $t$ that correspond to $r$}
                    \State $s \gets e_i$                                      \Comment{Use $e_i$ as subject $s$}
                    \State $o \gets e_j$                                      \Comment{Use $e_j$ as object $o$}
                    \State $d \gets (s,p,o,E,t)$                              \Comment{Group the alignments into $d$}
                    \State $D = D \cup \{d\}$                                 \Comment{Record the sample $d$}
                \EndIf
                \EndFor
            \EndFor
        \EndFor
        \Return $D_{\rm d}$, $D_{\rm ssm}$
        
        \item[] 
        \Procedure{PredicateMatcher}{$t,r$}
            \ForAll{alias string $a$ for $r$ in $K$} \Comment{WikiData holds a alias string collection for every predicate}
                \ForAll{substring $s$ in $t$}
                    \If{edit distance between $a$ and $s$ < 2}
                        \State \textbf{Return} $s$
                    \EndIf
                \EndFor
            \EndFor
        \EndProcedure
    \end{algorithmic}
\end{algorithm*}

Procedure~\ref{proc:data_construction} shows how we construct the pre-training data, including entity linking, predicate matching, triplet aligning, and deterministic relationship checking.
Each text piece $t$ is a paragraph in Wikipedia. We use the entity linker provided in \cite{entity-linker-REL}, represented as \textsc{EntityLinker}, to identify all the entities in the paragraph\footnotemark. The WikiData defines 12,043 aliases for 8,529 predicates. Function \textsc{PredicateMatcher} searches the substring corresponding to a predicate by comparing the predicate's aliases with all the substrings in the text. The identified predicate span is the nearest match whose edit distance is less than two. 
\footnotetext{{https://github.com/informagi/REL}}

After recognizing the entities and predicates, we combine every entity pair with every predicate as a triplet (entity, predicate, entity), enumerate all the possible combinations, and keep the ones that existed in KB as the triplets aligned with the paragraph. Line 7 checks if the object is deterministic by querying KB.
We record the obtained deterministic sample in the format of (subject $s$, predicate $p$, object $o$, text $t$, and entities $E$). 

The baselines use the pre-training sample as the following:
\begin{itemize}
[itemsep=1pt,topsep=1pt,parsep=2pt]
    \item \textbf{Deterministic} (mask deterministic object to train MLM): Get a sample $(s,p,o,E,t)$ from $D_{\rm d}$, mask the span corresponding to $o$ and train PLMs to predict $o$ based on the remaining context.
    \item \textbf{Random} (mask tokens randomly): Get a sample from $D_{\rm d}$, tokenize $t$ and calculate the number of tokens in $o$, denoted as $ \textsc{TokenCount}(o)$, randomly sample $\textsc{TokenCount}(o)$ tokens to be masked in the MLM training.
    \item \textbf{Whole word} (mask whole words randomly): Get a sample from $D_{\rm d}$, calculate the number of words in $o$ (separated by space), denoted as $ \textsc{WordCount}(o)$, randomly mask $\textsc{WordCount}(o)$ words in $t$.
    \item \textbf{Salient Span} (mask entities randomly): Get a sample $(s,p,o,E,t)$ from $D_{\rm ssm}$, randomly mask an aligned entity in $E$.
\end{itemize}
Although the masking granularity is different in the baslines, we keep the length of the masked content as similar as possible for a fair comparison. 

Then we introduce how the two proposed pre-training tasks use the data. In the \taska, the $s$ and $p$ are the deterministic clues and masked in the contrastive sample. If the same $o$ have more than one deterministic clue in $t$, e.g., multiple deterministic clues for the same $o$ are given by different triplets, all the deterministic $s$ and $p$ are considered as determined clues and masked in the contrastive sample \ref{sample:b}. In the \taskb, the number of the randomly masked token in the sample \ref{sample:c} is the same as the constrastive sample.

Procedure~\ref{proc:data_construction} is used to generate the \textbf{full data} (summarized in Table~\ref{tab:pre-training-data}). We obtain the \textbf{partial data} similarly, except that we do not need the \textsc{EntityLinker} (at Line 3 in Procedure~\ref{proc:data_construction}) and directly use the entity-text alignments provided in TREX.

\section{Hyperparameters}
\noindent\textbf{Pre-training} For the baselines trained on the partial data, the batch size is set to 512, the learning rate is $3\times 10^{-5}$, and the number of total training steps is 50,000. There are 200,000 training steps for the final model on the full data.

\noindent\textbf{Extractive QA}
\label{sec:app_hp}
In the experiments on extractive QA, we find that the model's performance is sensitive to hyperparameters. We conduct a grid search over the learning rate and batch size. In the separate setting, the learning rate is searched over $\left\{1\times 10^{-5}, 2\times 10^{-5}, 3\times 10^{-5}\right\}$, the batch size is searched over $\left\{16, 32, 64\right\}$ when fine-tuning every PLM on every dataset, and the epoch is set to 4. In the combine setting, the learning rate $\in \left\{2\times 10^{-5}, 3\times 10^{-5}\right\}$, the batch size $\in \left\{32, 64\right\}$, and the epoch is set to 2. We save the model checkpoint per 5,000 steps. The best model is selected from evaluating all the checkpoints. We use the code released by \cite{mrqa-code-source}\footnotemark.
\footnotetext{https://github.com/oriram/splinter}

\noindent\textbf{Closed-book QA}
The learning rate is set to $5\times 10^{-5}$. The training steps is set to 100,000 for NaturalQuestions and TriviaQA, 40,000 for WebQuestions. The model is evaluated per 10,000 training steps to select the best checkpoint.


\end{document}